\documentclass[runningheads]{llncs}
\usepackage[T1]{fontenc}
\usepackage{graphicx}
\usepackage[nolist]{acronym}
\usepackage{amsmath,amssymb,amsfonts}%
\usepackage{xcolor}%
\usepackage{booktabs}%
\usepackage{hyperref}%
\usepackage{xspace} 
\usepackage{subcaption}
\usepackage[labelformat=parens,labelsep=quad, skip=3pt]{caption}
\usepackage{siunitx}
\usepackage{orcidlink}  
\usepackage{comment}
\usepackage{array}

\begin{acronym}
\acro{AI}{Artificial Intelligence}
\acro{ANN}{Artificial Neural Networks}
\acro{AUC}{Area Under Curve }
\acro{CALD}{culturally and linguistically diverse}
\acro{CEDRIC}{The Comprehensive Epidemiological Database for Research, Innovation and Collaboration}
\acro{CNN}{Convolutional Neural Networks}
\acro{ED}{Emergency Department}
\acro{EHR}{Electronic Health Record}
\acro{ML}{Machine Learning}
\acro{NLP}{Natural Language Processing}
\acro{LLM}{Large Language Model}
\acro{ROC}{Receiver Operating Characteristic}
\acro{SDOH}{Social Determinants of Health}
\acro{SWERI}{South Western Emergency Research Institute}
\acro{SWSLHD}{South Western Sydney Local Health District}
\acro{MIMIC}{Medical Information Mart for Intensive Care}
\acro{NSW}{New South Wales}
\acro{BERT}{Bidirectional Encoder Representations from Transformers}
\acro{NIH}{National Institutes of Health}
\acro{BCB}{Bio-Clinical BERT} 
\acro{CH}{Classification Head}
\acro{NN}{Neural Network}
\acro{NN1}{Neural Network 1}
\acro{NN2}{Neural Network 2}
\acro{NN3}{Neural Network 3}
\acro{DS1}{Dataset 1: MIMIC-III Train Dataset}
\acro{HPC}{High Performance Computing}
\acro{MVA}{Motor Vehicle Accident}
\acro{BCBC}{Bio-Clinical BERT Classifier}
\acro{REGEX}{Regular Expression}
\acro{HREC}{Human Research Ethics Committee}
\acro{SM}{Supplementary Material}
\end{acronym}

\newcommand{\mimic}{\ac{MIMIC}-III\xspace}
\newcommand{\cedric}{\ac{CEDRIC}\xspace}   

\newcommand{\mimicData}{MIMIC data\xspace}   
\newcommand{\inghamOne}{CEDRIC One\xspace}   
\newcommand{\inghamTwo}{CEDRIC Two\xspace}  

\newcommand{\figwidtht}{0.32\textwidth}

\newcolumntype{C}[1]{>{\centering\arraybackslash}p{#1}}

\begin{document}
\title{Classification of kinetic-related injury in hospital triage data using NLP}
%
%
\author{Midhun Shyam\inst{1}\orcidlink{0009-0006-2930-2135} \and
Jim Basilakis\inst{1,2}\orcidlink{0000-0002-7440-1320} \and
Kieran Luken\inst{1}\orcidlink{0000-0002-6147-693X} \and
Steven Thomas\inst{2}\orcidlink{0000-0002-2416-0020} \and
John Crozier\inst{3}\orcidlink{0000-0002-4773-8518} \and
Paul M. Middleton\inst{2}\orcidlink{0000-0003-0760-1098} \and
X. Rosalind Wang\inst{1,2}\orcidlink{0000-0001-5454-6197}
}
\authorrunning{M. Shyam et al.}
%
\institute{School of Computer, Data and Mathematical Sciences, Western Sydney University, Victoria Rd, 2116, NSW, Australia \\
\email{\{J.Basilakis, rosalind.wang\}@westernsydney.edu.au} \\
\and
South Western Emergency Research Institute, Ingham Institute of Applied Medical Research, 1 Campbell St, Liverpool, 2170, NSW, Australia
\and
Liverpool Hospital, Liverpool, 2170 NSW, Australia
}

\maketitle              
\begin{abstract}

Triage notes, created at the start of a patient’s hospital visit, contain a wealth of information that can help medical staff and researchers understand \acl{ED} patient epidemiology and the degree of time-dependent illness or injury. Unfortunately, applying modern \acl{NLP} and \acl{ML} techniques to analyse triage data faces some challenges: Firstly, hospital data contains highly sensitive information that is subject to privacy regulation thus need to be analysed on site; Secondly, most hospitals and medical facilities lack the necessary hardware to fine-tune a \ac{LLM}, much less training one from scratch; Lastly, to identify the records of interest, expert inputs are needed to manually label the datasets, 
which can be time-consuming and costly. We present in this paper a pipeline that enables the classification of triage data using \ac{LLM} and limited compute resources. We first fine-tuned a pre-trained \ac{LLM} with a classifier using a small (2k) open sourced dataset on a GPU; and then further fine-tuned the model with a hospital specific dataset of 1000 samples on a CPU. We demonstrated that by carefully curating the datasets and leveraging existing models and open sourced data, we can successfully classify triage data with limited compute resources. 



\end{abstract}
\acresetall      

\section{Introduction}
\label{sec:intro}

Triage is the process of assessing patients, and assigning a level of urgency to be seen by medical staff. Triage notes are free-text clinical documentation created by a triage nurse, at the start of a patient's visit to an \ac{ED}. They contain a short paragraph reflecting the nurse's assessment of the presenting complaint and the condition of the patient; thus they contain vital information about a patient's visit and risk stratification of their needs. Moreover, triage notes are completed for all ED patients, representing more than half the clinical cases that a hospital sees. 

This wealth of information can help us understand ED patient epidemiology and the degree of time-dependent illness or injury, as well as the underlying characteristics of the general population.

In this study, we aimed to apply modern \ac{NLP} \acp{LLM} with \ac{ML} classification to classify triage data to identify patients who presented to hospital with a specific complaint. Specifically, we were interested in patients who had a kinetic energy related vehicular trauma, such as injury within a vehicle, collision with and falls from a moving vehicle. Results from the study could inform policy decisions such as road safety legislation that aims to achieve zero road-related fatalities and serious injuries.

A recent review~\cite{stewart2023plosone} of the application of \ac{NLP} on triage records identified 20 publications, most of which applied \ac{NLP} to free text notes to predict outcomes that exist within the structured data associated with each patient. 
For our work however, we required labels that were not a part of the \ac{EHR} in any form. 

The primary objective of this study was to develop and implement a highly accurate context-based text classification model for categorising triage notes, which are contained in a linkage and analytic healthcare platform known as \cedric database~\footnote{Maintained by \ac{SWERI} and represents \ac{EHR} data derived from the \ac{SWSLHD} hospitals.}. 
Specifically, we aimed to classify emergency cases into kinetic-related accident cases and non-kinetic-related accident cases. This is often not straightforward; for example, an incident where a seizure or heart attack occurs while driving 
is more accurately classified as a case of seizure or heart attack rather than a general trauma \acl{MVA} case. Therefore, accurate classification of these cases demonstrates the ability of models to understand semantics. 

To this end, we adopted a BERT model~\cite{devlin2019bert} that was trained specifically with clinical data --- \ac{BCB}~\cite{alsentzer2019publicly} --- for the purpose of classifying triage data. 
Using a pre-trained model significantly reduced the computational resources needed for our task. Importantly, triage notes differ from most medical text data in their use of jargon, acronyms, and shorthand. Despite these specialised aspects, we demonstrate that by fine-tuning the \ac{BCB} model with small manually labelled datasets for our specific purpose, we could still achieve good results. 
To our knowledge, this is the first work to classify triage data by fine-tuning a pre-trained model with a small manually labelled dataset.

\section{Methods}
\label{sec:methods}

\subsection{Data}

\subsubsection{\ac{MIMIC}-III data}

The \mimic Clinical Database~\cite{johnson2016mimic3db,goldberger2000physiobank} comprises observations with presenting problems of over 40,000 critical care patients between 2001 and 2012. 
We used the table \texttt{NOTEEVENTS} that contains various de-identified clinical notes, for this paper. We created from this table a dataset --- henceforth called \emph{\mimicData} --- containing 1309 positive kinetic-related vehicular injury cases and 1132 negative cases. Full description of the data pre-processing is in the \ac{SM} in the project GitHub repository.

\subsubsection{\cedric data}

\ac{CEDRIC} database captures and links data from \ac{ED} presentations in \ac{SWSLHD} since 2005. For this work, we concentrated on the \ac{ED} triage comments in \cedric, and created two data sets (full details in \ac{SM}): 
\begin{enumerate}
    \item \inghamOne : positive cases from 2022 and both cases from 2021, resulting in 447 kinetic injury cases and 553 others. 
    \item \inghamTwo : positive cases from 2023 and both cases from 2022, resulting in 413 kinetic injury cases and 587 others. 
\end{enumerate}

\subsubsection{Ethics}

The \mimic data is publicly available and does not require Ethics approval. The \cedric data is under the approval for the Multicultural Emergency Medicine Epidemiology (MEME) study (HREC Reference: LNR/17/LPOOL/432; SSA Reference: LNRSSA/17/LPOOL/433; Project number: HE17/234)

\subsubsection{Data Availability}

The labelled \mimicData is available in the GitHub repository for this work. The \cedric data sets used are not available to the public due to ethics considerations and privacy regulations.

\subsection{Deep Learning Classifier}

In this work, we leveraged the pre-trained \ac{BCB} model~\cite{alsentzer2019publicly} as a base for the \ac{NLP} embedding, due to the model's small size (100 million parameters) and computational efficiency. 
For our downstream classification task, we appended a two‑dimensional sequence \ac{CH} to \ac{BCB}'s encoder for binary output. 
This model was referred to as \emph{\ac{BCBC}}, and was fine-tuned in three different configurations: 
\begin{enumerate}
    \item \textbf{\acf{NN1}} --- The weights from the \ac{BCB} were used \emph{as is} (frozen) and the fine-tuning was used to optimise the \ac{CH} parameters only. 
    \item \textbf{\acf{NN2}} --- All but the last layer of the \ac{BCB} were frozen, and the fine-tuning process was used to optimise the weights of the \ac{CH} and the final encoder layer (layer 12). 
    \item \textbf{\acf{NN3}} --- Fine-tuning to optimise the weights for layers 11 and 12 of the \ac{BCB} and the \ac{CH}. 
\end{enumerate}

We experimented with Adam, AdamW, and SGD optimisers for learning, and different values of dropout rates ($dr = [0.15, 0.20, 0.25]$) and learning rates ($lr = [0.0001, 0.0005, 0.005]$) to find the optimal hyperparameter.

All permutations of network configuration, optimiser and hyperparameters were run ten times to account for variability due to the stochastic nature of \acp{NN}.

\subsection{Experimental Procedure}

We carried out the experiment in the following steps: 
\begin{enumerate}
    \item Fine-tuning \ac{BCBC} with all 81 permutations of network configuration, optimiser and hyperparameters using \mimicData. 
    \item Prediction of \inghamTwo data using the fine-tuned models. 
    \item Further fine-tune (domain adaptation) the models from Step 1 using \inghamOne data. 
    \item Final prediction of \inghamTwo data using the domain adapted models from Step 3. 
\end{enumerate}
For both fine-tuning steps, 80\% of the data was used for training and the rest for validation. Further, they were set to run for 200 epochs with an early stopping trigger if the validation loss did not improve for ten epochs. 

The two-step fine-tuning is implemented for one reason --- data from hospitals contain a lot of private information and cannot be relocated off-site to train models using high-performance computing resources. While fine-tuning a pre-trained model uses a lot fewer resources than fully training a \ac{LLM}, it could still require the use of a modern GPU. This is a resource that is often not available at a medical facility. Therefore, instead of fine-tuning the classifier with just project-specific data, we fine-tuned with a publicly available data set. This allows other medical researchers to perform the same task on cloud-based GPU facilities.

\subsection{Performance metrics}

We evaluated the performance of the classifier using the following metrics: accuracy, F1-score and total time taken. We also calculated and analysed the confusion matrix entities (true/false positive and negative values), as well as precision and recall for the prediction results. However, we omit these results here as they exhibit a similarity in trend with accuracy and F1-scores.

\subsection{Software and Hardware}

All \acp{NN} were fine-tuned with \mimicData using the \ac{HPC} resources at Western Sydney University, 
on NVIDIA A100 GPUs with 32GB of RAM and 40GB of GPU memory. 
Further fine-tuning (domain adaptation) and all predictions were performed on Intel Xeon Gold 6242R CPU @ 3.10GHz using one CPU core and 4GB RAM. 

The \ac{BCB} model is available from Hugging Face~\footnote{\url{https://huggingface.co/emilyalsentzer/Bio_ClinicalBERT}}. All code, including significance tests, developed in this work is available on GitHub~\footnote{\url{https://github.com/CRMDS/Kinetic-Injury-Triage}}.

\section{Results and Discussion}
\label{sec:results}

We present here summaries of the results, full results, and \ac{SM} with all relevant plots, tables and discussions are in the project GitHub repository. 

\subsection{Fine-tuning on \mimicData}
\label{sec:finetune_mimic}

We found that most learning/dropout rate combinations provided similar 
results, with the exception of the \ac{NN3} configuration using $lr = 0.005$. The \ac{NN3} architecture with Adam ($lr = 0.0001$, $dr = 0.15$) achieved the best metrics (accuracy: $95.0 \pm 0.9\%$, F1-score: $95.4 \pm 0.8\%$), but trained slower than AdamW, which was up to three times faster with only slightly lower performance.

A two-sample t-test ($N=10$) across all results showed about half of the pairs had significant differences ($p = 0.05$), though many were marginal. 
However, the difference of $\sim 1\%$ between the best and the worst results lacks practical impact even though they were statistically significantly different. 

\subsection{Prediction with fine-tuned models}
\label{sec:prediction1}

We found the prediction results from the fine-tuned models mirror those in Section~\ref{sec:finetune_mimic}, with the Adam and AdamW optimised models providing roughly the same accuracies and F1-scores. Like Section~\ref{sec:finetune_mimic}, the SGD optimised models, and $lr = 0.005$ perform the worst. While about half of the result pairs exhibited a statistically significantly difference, the practical difference for clinicians is minimal. 
We note that the drop in accuracy from $\sim95\%$ reported Section~\ref{sec:finetune_mimic} to $\sim84\%$ is not unexpected, as the \mimicData and the \inghamTwo data have fundamentally different context. 

The total prediction time was similar across all configurations -- between 170 and 290 seconds to predict the data set of 1000 triage notes. Hence, this task could easily be achieved on much larger datasets within medical facilities. 

Given the significant under-performance of the \ac{NN1} architecture, the SGD optimised models, and the models with a learning rate of 0.005, we did not continue further testing with these models. 

\subsection{Further fine-tuning on \inghamOne Data}
\label{sec:domain_adaptation}

Similar to Sections~\ref{sec:finetune_mimic} and \ref{sec:prediction1}, 
we found the Adam optimised models provided the best overall performance with the best accuracy (using $lr = 0.0005, dr = 0.2$, gave an accuracy of $94.5 \pm 0.9 \%$) and F1-score (using $lr = 0.0001, dr = 0.15$, and gave an F1-score of $94.0 \pm 1.2\%$). We note that this configuration provided the equal best average accuracy, however had a larger standard deviation than the other model identified). 

Unlike the previous sections, all pairs of results reported were \emph{not} statistically significantly different from each other (at $p=0.05$). Once again, however, the practical difference between an accuracy of 94.5\% (the best performing model overall) and 93.5\% (the worst performing model) is not significant.

\subsection{Prediction with domain adapted models}
\label{sec:prediction2}

\begin{figure}[tb]
	\begin{center}
		\begin{subfigure}[b]{\figwidtht}
			\caption{} 
			\includegraphics[width=\textwidth]{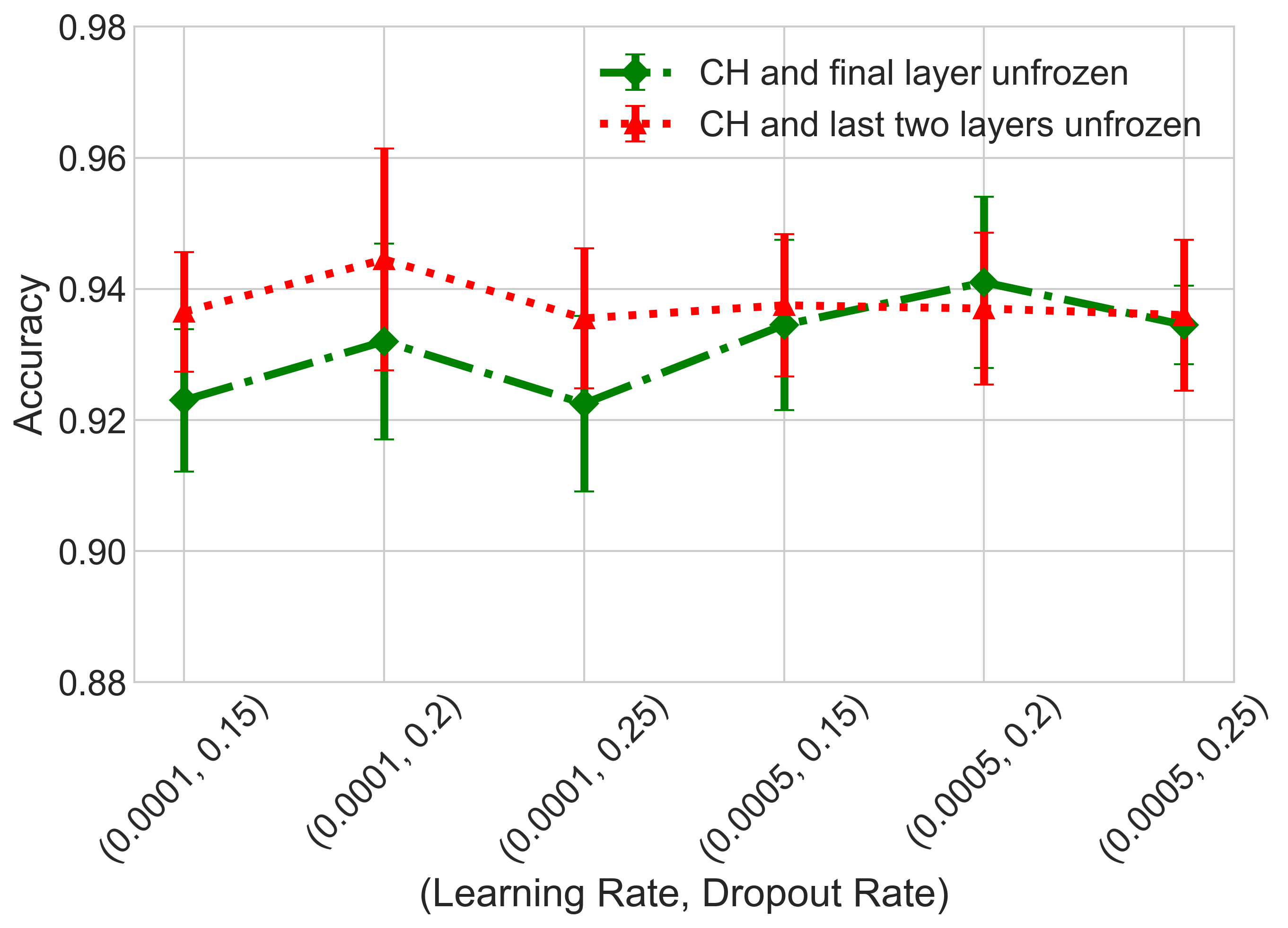}
		\end{subfigure}
        \hfill
		\begin{subfigure}[b]{\figwidtht}
			\caption{}
			\includegraphics[width=\textwidth]{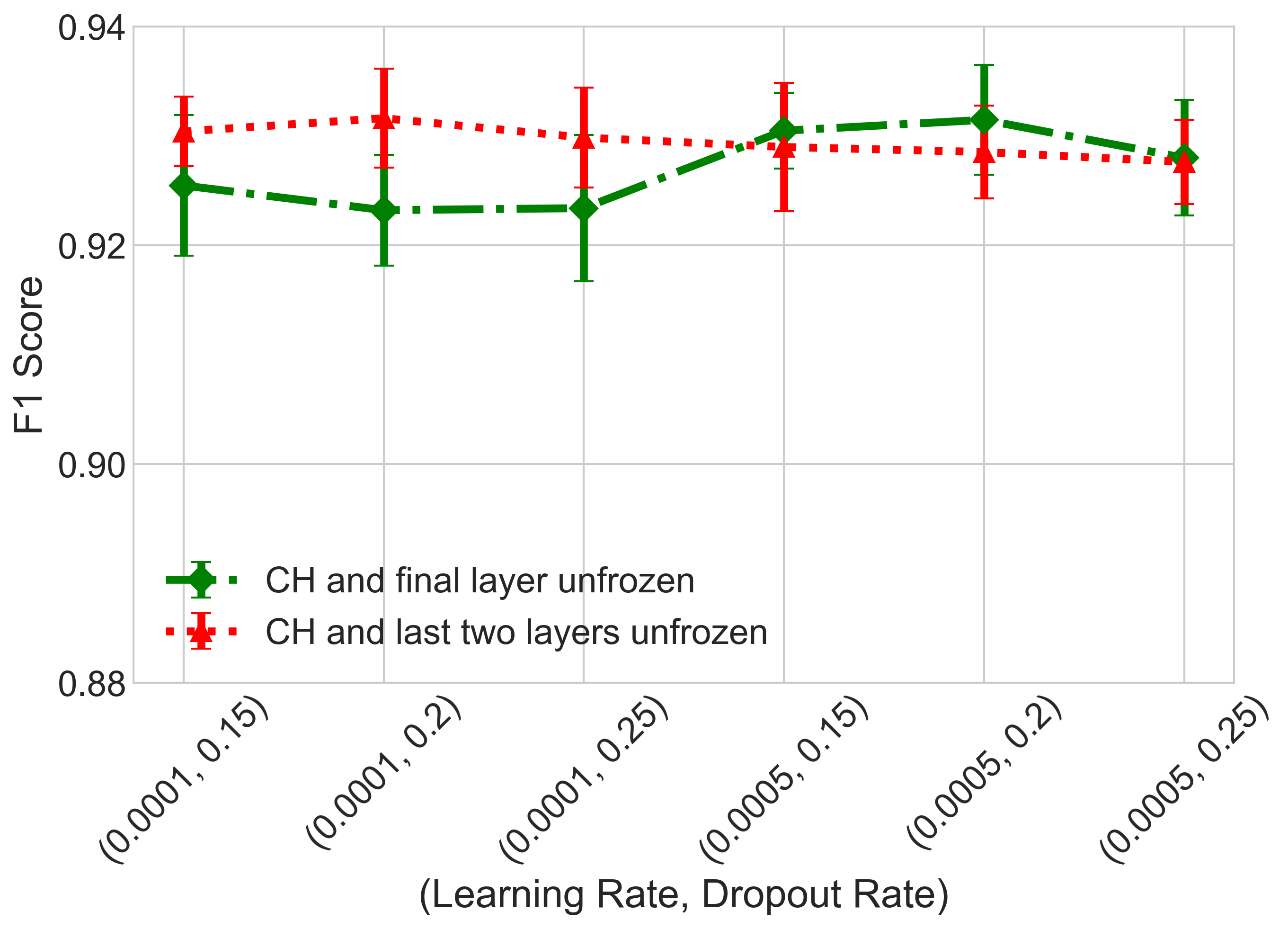}
		\end{subfigure}
        \hfill
		\begin{subfigure}[b]{\figwidtht}
			\caption{}
			\includegraphics[width=\textwidth]{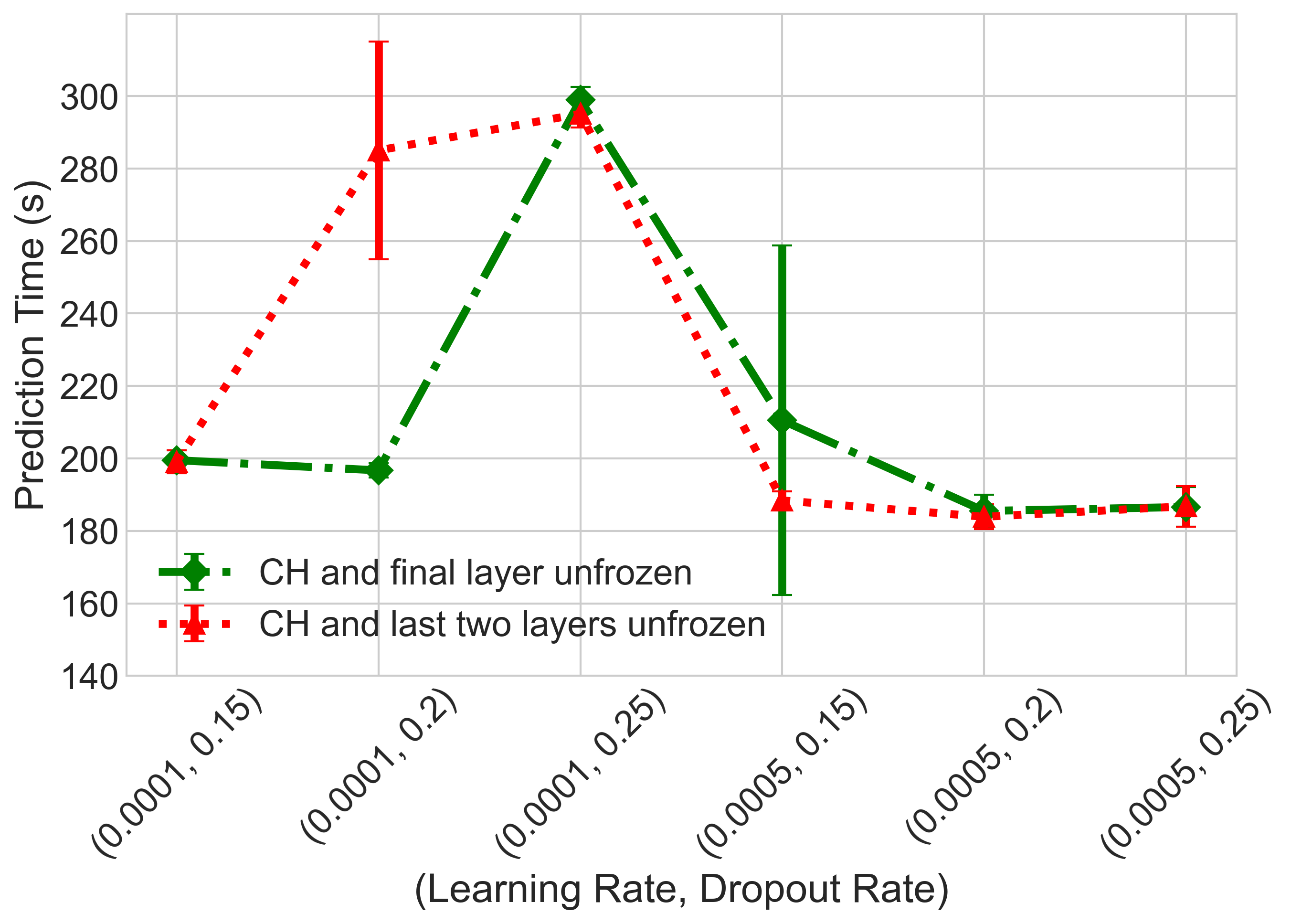}
		\end{subfigure}
	\end{center}
	\caption{Results of prediction on \inghamTwo data with the domain adapted models optimised with AdamW optimiser: (a) Prediction accuracy; (b) F-1 score; (c) Time taken on CPU (s)
	} 
	\label{fig:res_predict2}
\end{figure}

\begin{table}[tb]
\caption{Accuracy and F1-score of prediction on \inghamTwo data using the domain adapted \ac{NN3} models. 
The best result for each optimiser is highlighted. }
\label{tab:res_prediction2}
\centering
	\resizebox{\textwidth}{!}{%
    \begin{tabular}{@{}lcccccc@{}}
    \toprule
    &  \multicolumn{6}{c}{Accuracy (Learning rate, Drop out rate)} \\ \cmidrule(l){2-7} 
Optimiser & (0.0001, 0.15) & (0.0001, 0.2) & (0.0001, 0.25) & (0.0005, 0.15) & (0.0005, 0.2) & (0.0005, 0.25) \\ \midrule
Adam & $0.931 \pm 0.004$ & \textbf{0.935 $\pm$ 0.004} & $0.932 \pm 0.004$ & $0.933 \pm 0.005$ & $0.934 \pm 0.006$ & $0.932 \pm 0.005$ \\
AdamW & $0.930 \pm 0.003$ & \textbf{0.931 $\pm$ 0.004} & $0.929 \pm 0.004$ & $0.929 \pm 0.006$ & $0.928 \pm 0.004$ & $0.927 \pm 0.004$ \\
    \bottomrule
        \end{tabular}
    }
\vspace{2mm}

	\resizebox{\textwidth}{!}{%
    \begin{tabular}{@{}lcccccc@{}}
    \toprule
    &  \multicolumn{6}{c}{F1 Score (Learning rate, Drop out rate)} \\ \cmidrule(l){2-7} 
Optimiser & (0.0001, 0.15) & (0.0001, 0.2) & (0.0001, 0.25) & (0.0005, 0.15) & (0.0005, 0.2) & (0.0005, 0.25) \\ \midrule
Adam & $0.920 \pm 0.004$ & \textbf{0.924 $\pm$ 0.004} & $0.921 \pm 0.004$ & $0.921 \pm 0.006$ & $0.922 \pm 0.007$ & $0.920 \pm 0.005$ \\
AdamW & $0.919 \pm 0.003$ & \textbf{0.920 $\pm$ 0.005} & $0.918 \pm 0.005$ & $0.916 \pm 0.006$ & $0.916 \pm 0.004$ & $0.915 \pm 0.004$ \\
    \bottomrule
        \end{tabular}
    }

\end{table}

The final prediction results using the domain adapted models from Section~\ref{sec:domain_adaptation} are presented in Figure~\ref{fig:res_predict2} and Table~\ref{tab:res_prediction2}. We found that the results aligned with the conclusions drawn from the previous sections. That is, the best models optimised with Adam and AdamW, using $lr = 0.0001, dr =0.2$, had near-parity in accuracies ($93.5 \pm 0.4\%$ and $93.1 \pm 0.4\%$) and F1-scores ($92.4 \pm 0.4\%$ and $92.0 \pm 0.5\%$). Unsurprisingly, the domain adapted \ac{NN3} models outperformed the fine-tuned models from Section~\ref{sec:finetune_mimic} when tested on the same data. 

Similar to the results in Section~\ref{sec:domain_adaptation}, for almost all pairs of results, the difference between the values were not statistically significant.

\subsection{Discussion}
\label{sec:discussion}

We fine-tuned 81 \ac{BCBC} models with varying parameters (optimisation methods, \ac{NN} fine-tuning configurations, learning rates, and drop out rates) using the \mimicData. These models were then tested on \inghamTwo data, demonstrating generalisability. Domain adaptation using \inghamOne data showed improvement of the classification accuracies on \inghamTwo data, showing the benefit of training models on data with similar context. 

Overall, Adam and AdamW optimisers performed similarly, outperforming the SGD optimiser. Similarly, fine-tuning only the \ac{CH} (keeping all pre-trained weights from \ac{BCB}) resulted in worse models than unfreezing either the last or the last two layers of the pre-trained \ac{BCB}. The latter two architectures showed comparable results. This was unsurprising as the pre-trained model was not trained for any task, and those weights were not optimised for classification purpose. Lastly, we tested three different values of learning and dropout rates each. We note that these were not the default values of these hyperparameters as set by the \texttt{pytorch} package that \ac{BCB} is built with. Of the nine different permutations of these parameters, only those with the highest learning rate ($lr = 0.005$) resulted in significantly worse accuracies than the others, showing there is a range of parameters that can achieve optimum or near optimum classification results; thus, there will be no need to test a large range of hyperparameters to adapt the method in practise. 

The total time taken to follow the learning procedure as described in this work is feasible for any group: The majority of the time for producing the result was taken up by manually labelling the various data sets. In this work, the labels were mainly produced by two of the authors who are experts in emergency and trauma medicine. We estimated that the labelling of each of the triage data sets of 1000 samples took somewhere between two and four hours. The pre-processing and labelling of the \mimic data was estimated to be about one day's work if using the code developed for this work. 

The fine-tuning using AdamW of \ac{NN3} took under ten minutes on one GPU, which is easily achievable on a cloud based compute service as as Google Colab, Kaggle, Hugging Face, etc.\footnote{We note that this step is also achievable without a GPU: AdamW using $lr = 0.0001, dr = 0.15$ took $4.94 \pm 0.94$ hours to finetune using 32 CPUs.} The second largest time spent for this work was the domain adaptation of the models using \inghamTwo data set, as this was done on a CPU with very limited resources. Despite this, the models were fine-tuned in between 1.5 and 3 hours, something that is feasible for anyone with a computer. 

The prediction for 1000 samples of the test data varied between 170 and 300 seconds. Our local health district receives around half a million patients in \ac{ED} in a year, which means that even with minimum computing resources (one CPU core and 4GB RAM) the classification process would only take approximately one day for a year's data. In reality, a modern computer should be able to classify half a million records within a few hours\footnote{We showed in Supplementary Materials that as we include the number of CPUs, the fine-tuning and the prediction time decreased dramatically.} . 

Overall, the work presented here showed that the pre-trained \ac{BCB} can easily be adapted for the classification of proprietary hospital triage data with clinically useful results. As patient data are under strict privacy laws, and most medical facilities do not possess high-end GPU compute resources, the procedures presented in this work provide a way forward for medical facilities to use modern \acp{LLM} to extract and classify the data held within their own patient records.

\section{Conclusion}

\acl{ED}s in hospitals sees the most undifferentiated and heterogeneous cohort of patients in the whole health system. Triage notes represent a patient's first encounter with the system and summarises a patient's current status, their past history, and how urgently they need to see a doctor (risk stratification). 
As such, the ability to analyse and classify triage notes would help health systems to understand their patient cohort, epidemiology, and degrees of illness or injury, and treat them appropriately and safely. The significance of our approach is the democratisation of the use of modern \ac{NLP}, \ac{LLM} and \ac{ML} technologies for every medical facility. The pipeline we presented would enable these facilities to classify the type of information they are seeking from their own clinical notes. 

We demonstrated that even with limited compute resources, we could achieve high classification accuracies in our use case of determining kinetic-related vehicular injuries. 
This work contributes a scalable and privacy-preserving clinical \ac{NLP} methodology that enables secure handling of sensitive data through local deployment, with an estimated end-to-end replication time of 3–4 days.

\begin{credits}
\subsubsection{\ackname}
MS wishes to thank Stuart Fitzpatrick for insightful discussions that improved early development of this work.
\subsubsection{\discintname}
The authors have no competing interests to declare that are
relevant to the content of this article. 

\end{credits}

%
%
%
\bibliographystyle{splncs04}
\bibliography{references}

\end{document}